\PassOptionsToPackage{numbers}{natbib}
\documentclass[conference]{IEEEtran}
\IEEEoverridecommandlockouts
\usepackage{cite}
\usepackage{amsmath,amssymb,amsfonts}
\usepackage{algorithmic}
\usepackage{graphicx}
\usepackage{textcomp}
\usepackage{stfloats}
\usepackage{float}
\usepackage{xcolor}
\usepackage{booktabs}
\usepackage{multirow}
\usepackage{url}
\usepackage{natbib}
\usepackage{amsmath,amssymb,bm}
\usepackage{tikz}
\usetikzlibrary{calc,positioning,arrows.meta,backgrounds,fit}
\usepackage{booktabs}
\usepackage{xcolor}
\usepackage{tikz}
\usetikzlibrary{arrows.meta, positioning, fit, backgrounds}

\usetikzlibrary{shapes,arrows,positioning,calc}
\usepackage{amsmath}
\usepackage{amssymb}
\usepackage{caption}
\usepackage{cuted}
\usepackage{placeins}

\def\BibTeX{{\rm B\kern-.05em{\sc i\kern-.025em b}\kern-.08em
    T\kern-.1667em\lower.7ex\hbox{E}\kern-.125emX}}
\begin{document}

\title{AlphaJet: Automated Conceptual Aircraft Synthesis via Disentangled
Generative Priors and Topology-Preserving Evolutionary Search
}

\author{
\IEEEauthorblockN{Boris Kriuk}
\IEEEauthorblockA{
    Hong Kong University of Science and Technology\\
    Clear Water Bay, Hong Kong, HKSAR\\
    bkriuk@connect.ust.hk}
}

\maketitle

\begin{abstract}
Conceptual aircraft design is traditionally an expert-mediated iterative
process in which a human designer proposes a configuration, runs low-order
physics, inspects the result, and re-proposes. We present AlphaJet, an
end-to-end automated synthesis pipeline that closes this loop. From a textual
mission specification (mass, range, cruise speed, hard size envelope, engine
count, areal density) AlphaJet evolves a feasible 3D aircraft in real time,
scored by a transparent multi-disciplinary fitness function covering
aerodynamics, structures, weights, stability, packaging, and geometric mount
consistency. Three contributions distinguish our approach: (i) an
Anatomically-Disentangled Variational Autoencoder (AD-VAE) whose first $25$
latent dimensions are supervised to align with named anatomical parameters,
providing an interpretable shape prior; (ii) a topology-elitist genetic
algorithm that protects the best individual from each of five tail topologies
and triggers stagnation restarts, preventing premature collapse to a single
configuration; and (iii) mount-aware geometric scoring that computes signed
penetration between engines and other structural parts,
eliminating the redundant artifacts common in generative aircraft models.
The full loop runs interactively on a CPU and streams every generation to a
browser viewer, making it a practical real-world automation tool for early-phase
design-space exploration.
\end{abstract}

\begin{IEEEkeywords}
design automation, generative models, evolutionary computation, aircraft
conceptual design, multidisciplinary optimization, variational autoencoder.
\end{IEEEkeywords}

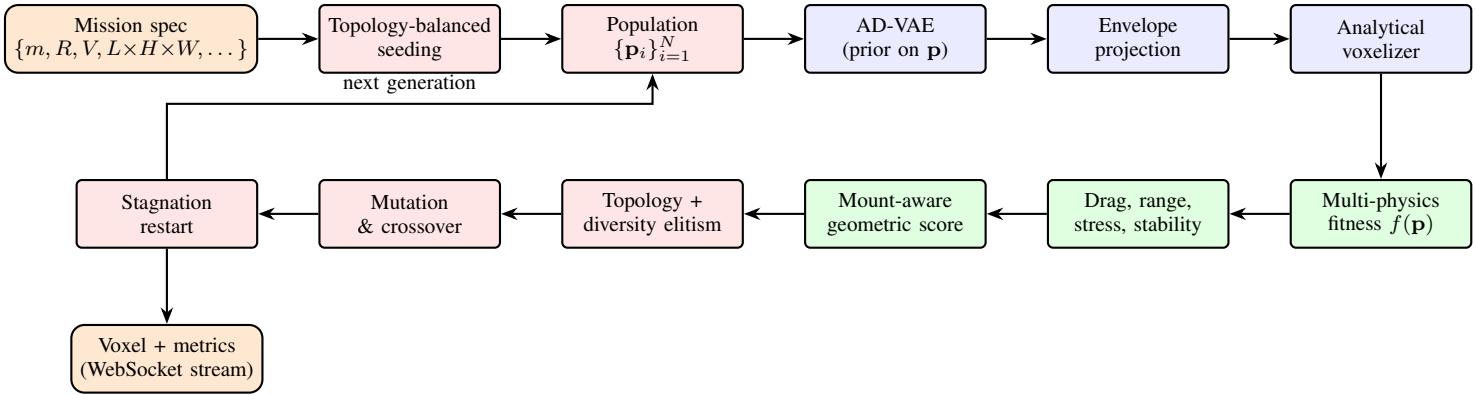
\begin{figure*}[!t]
\centering
\begin{tikzpicture}[
  font=\footnotesize,
  >={Stealth[length=2.2mm]},
  every path/.style={thick},
  block/.style={rectangle, rounded corners=2pt, draw, minimum height=9mm,
                minimum width=24mm, align=center, fill=blue!8},
  data/.style ={rectangle, rounded corners=5pt, draw, minimum height=9mm,
                minimum width=24mm, align=center, fill=orange!18},
  phys/.style ={rectangle, rounded corners=2pt, draw, minimum height=9mm,
                minimum width=24mm, align=center, fill=green!12},
  ga/.style   ={rectangle, rounded corners=2pt, draw, minimum height=9mm,
                minimum width=24mm, align=center, fill=red!10},
]

\node[data]                   (mission) {Mission spec\\$\{m,R,V,L{\times}H{\times}W,\dots\}$};
\node[ga,    right=8mm of mission] (seed)    {Topology-balanced\\seeding};
\node[ga,    right=8mm of seed]    (pop)     {Population\\$\{\mathbf{p}_i\}_{i=1}^{N}$};
\node[block, right=8mm of pop]     (advae)   {AD-VAE\\(prior on $\mathbf{p}$)};
\node[block, right=8mm of advae]   (project) {Envelope\\projection};
\node[block, right=8mm of project] (vox)     {Analytical\\voxelizer};

\node[phys,  below=14mm of vox]     (phys)    {Multi-physics\\fitness $f(\mathbf{p})$};
\node[phys,  left=8mm of phys]      (mdo)     {Drag, range,\\stress, stability};
\node[phys,  left=8mm of mdo]       (mount)   {Mount-aware\\geometric score};
\node[ga,    left=8mm of mount]     (select)  {Topology +\\diversity elitism};
\node[ga,    left=8mm of select]    (mut)     {Mutation\\\& crossover};
\node[ga,    left=8mm of mut]       (restart) {Stagnation\\restart};

\node[data, below=10mm of restart] (out) {Voxel + metrics\\(WebSocket stream)};

\draw[->] (mission) -- (seed);
\draw[->] (seed)    -- (pop);
\draw[->] (pop)     -- (advae);
\draw[->] (advae)   -- (project);
\draw[->] (project) -- (vox);

\draw[->] (vox.south) -- (phys.north);

\draw[->] (phys)   -- (mdo);
\draw[->] (mdo)    -- (mount);
\draw[->] (mount)  -- (select);
\draw[->] (select) -- (mut);
\draw[->] (mut)    -- (restart);

\draw[->] (restart.north) |- ($(pop.south)+(0,-4mm)$)
          node[pos=0.75, above] {next generation} -- (pop.south);

\draw[->] (restart.south) -- (out.north);

\end{tikzpicture}
\caption{AlphaJet architecture. The genetic algorithm outer loop (red blocks)
maintains a population of anatomical vectors. Each candidate is regularized
by the AD-VAE prior, projected onto the user envelope, and voxelized
analytically (blue blocks). The voxel grid is scored by a transparent
multi-physics evaluator (green blocks), and the best individual of every
generation is streamed to a browser viewer.}
\label{fig:arch}
\end{figure*}

\section{Introduction}

Conceptual-phase early-stage aircraft design sits between two automation gaps. On one side, multidisciplinary design optimization (MDO) frameworks automate
\emph{evaluation} of a fixed configuration but leave \emph{shape proposal} to
a human. On the other, generative deep models automate \emph{shape proposal}
but produce geometry without physical commitments, frequently emitting
configurations whose engines float in space, whose tails do not touch a fin,
or whose volumes cannot close on fuel and payload. Bridging the two is hard
because the search space is mixed-topology (tail type, engine count, mount
location are categorical) and feasibility is governed by interacting
constraints (range, stability, structure, packaging) that are cheap
individually but punishing in combination.

The economic stakes of conceptual design are high \cite{mohaghegh2025machine, gao2020filter}. Decisions taken in the
first weeks of a program (engine count, wing aspect ratio, tail topology,
overall dimensions) lock in roughly $80\%$ of the lifetime cost of the
aircraft, even though they account for less than $5\%$ of the engineering
hours expended over the program. Mistakes made at this stage are
prohibitively expensive to unwind once detailed CAD, certification analyses,
and supplier contracts are anchored to a baseline \cite{diedrich2006multidisciplinary}. An automation tool that
can quickly explore the feasible region of the configuration space, surface
non-obvious topology trade-offs, and reject inconsistent geometry before a
human invests effort in it therefore has direct industrial value, even when
its underlying physics is deliberately first-order \cite{sobester2014aircraft, sabater2022fast}.

We introduce AlphaJet, an automated ML framework that addresses such gap.
A user supplies a mission and a hard bounding box; AlphaJet evolves a
feasible voxel-rendered aircraft in minutes on a CPU. The pipeline
(Fig.~\ref{fig:arch}) combines a learned shape prior with an evolutionary
outer loop and a closed-form physics evaluator. Crucially, the neural
network is used as a regularizer of the search space, not as the geometry
generator: every candidate is rendered analytically, which prevents neural
hallucination from contaminating fitness. Such single architectural
commitment, more than any individual algorithmic detail, is what makes the
emitted designs auditable: every voxel can be traced back to a named
anatomical parameter and every fitness contribution can be traced back to a
closed-form physical expression.

The contributions of this work are summarized as follows.
\begin{itemize}
\item An Anatomically-Disentangled VAE (AD-VAE) whose latent disentanglement
is enforced by construction via a supervised alignment loss and a split KL
divergence, rather than discovered post-hoc.
\item A topology-preserving genetic algorithm with per-topology elitism,
diversity elitism, and stagnation restarts, which automatically maintains
structural-class coverage without human curation.
\item Mount-aware geometric scoring, a continuous signed-penetration metric
that penalizes parts which fail to physically attach to their host
structure, producing generative output that is geometrically consistent.
\item An interactive automation tool: the full loop streams every
generation's voxel cloud and $30{+}$ sub-metrics to a browser viewer in real
time.
\end{itemize}

\section{Related Work}

AlphaJet sits between three communities. MDO frameworks \cite{ciampa2020agile, perez2004evaluation} such as
SUAVE, OpenMDAO, and OpenVSP automate physics evaluation and gradient-based
refinement but require a human to seed the topology. They typically expose a
parametric model (a wing, a fuselage, a tail) with continuous degrees of
freedom and rely on a domain expert to choose the categorical structure
beforehand; once that choice is fixed the optimizer cannot escape it,
which is precisely the bottleneck early-phase studies are trying to relieve.
Generative shape models (3D GANs, voxel VAEs, signed-distance networks
trained on aircraft corpora) \cite{chen2019learning, nash2017shape, chaudhuri2020learning, kriuk2026pstnet} automate shape proposal but couple weakly,
if at all, to physics, so their outputs satisfy visual plausibility rather
than mission feasibility. They also tend to inherit the biases of the
training set --- a corpus dominated by tube-and-wing airliners produces a
prior that is reluctant to emit blended-wing-body or canard configurations
even when the mission would favor them. Evolutionary aircraft design
couples search to physics but tends to collapse onto a single tail topology
once one configuration begins to dominate, requiring manual restarts \cite{kriuk2025shepherd, tfaily2024bayesian}. A
parallel line of work in novelty search and quality-diversity
algorithms explicitly maintains coverage of a behavioral archive, but
behavior characterizations for aircraft are not well standardized and the
archives can balloon in size for problems with many categorical axes \cite{kriuk2025gelovec, li2023intelligent, bartoli2016improvement}.
AlphaJet's contribution is a closed loop that inherits the search-space
regularization of generative priors, the configuration coverage of
multi-niche evolutionary algorithms, and the auditability of low-order
analytical physics, while keeping the diversity-preservation mechanism
small enough to run on a single CPU.

\section{Method}

\subsection{Anatomical Parameterization}

Each aircraft is described by a $25$-dimensional anatomical vector
$\mathbf{p} \in \mathbb{R}^{25}$ covering fuselage geometry (length, radius,
nose and tail fineness), main wing (span, root chord, taper, sweep, dihedral,
chordwise and vertical position, thickness), empennage (vertical tail size,
sweep, cant; horizontal tail span, chord, vertical position; binary
existence flags), and propulsion (engine count, length, size, longitudinal
position, spanwise spread). Each parameter has an explicit physical range and
is normalized to $[-1,1]$ before being passed to either the VAE or the GA.
The anatomical vector decodes deterministically through an analytical
voxelizer that handles the asymmetric fuselage profile (blunt cockpit nose,
sharper upswept tail cone), yehudi-break wings, five tail topologies, and
either fuselage-mounted or wing-podded engines. The choice of $25$ scalars
is deliberately conservative: it is rich enough to span the configurations
encountered in a wide spectrum of subsonic transports, business jets, and
medium-altitude long-endurance UAVs, but sparse enough that the GA can
explore it densely with a population of order $10^2$ individuals. We found
that pushing the dimensionality higher (e.g. by adding airfoil-shape
coefficients per spanwise station) produced visually richer geometry but
diluted the signal in every individual fitness component, slowing
convergence without measurably improving the final design.

\subsection{Architecture Overview}

The pipeline (Fig.~\ref{fig:arch}) is a three-block control loop. A genetic
algorithm maintains a population of normalized anatomical vectors. Each
candidate is passed through the AD-VAE prior (which keeps the anatomy on a
shape-coherent manifold learned from $4{,}000$ synthetic jets), projected
onto the user-supplied bounding envelope, and voxelized analytically. The
resulting $384^3$ voxel grid feeds the multi-physics evaluator, which returns
a scalar fitness and a $30$-element diagnostic breakdown. Selection,
mutation, topology elitism, and stagnation restarts complete the loop.

\subsection{Anatomically-Disentangled VAE}

Standard $\beta$-VAEs achieve disentanglement by tuning a regularization
coefficient and inspecting the result. We instead enforce disentanglement
by construction: the first $25$ latent dimensions of a $48$-D latent
space are supervised to match the ground-truth anatomical vector, while the
remaining $23$ free dimensions absorb residual shape variation under
standard KL pressure. The training objective for an input voxel grid
$\mathbf{x}$ with ground-truth normalized parameters $\mathbf{p}^{\star}$ is
\begin{equation}
\label{eq:advae}
\begin{aligned}
\mathcal{L}_{\text{AD-VAE}} \;=\;
& \underbrace{\mathrm{BCE}(\hat{\mathbf{x}}, \mathbf{x})}_{\text{recon}} \;+\;
  \beta \!\left(\alpha\, D_{\mathrm{KL}}^{\text{anat}} \!+\! D_{\mathrm{KL}}^{\text{free}}\right) \\
& +\; \lambda_{\text{anat}}\,\bigl\|\boldsymbol{\mu}_{1{:}25} - \mathbf{p}^{\star}\bigr\|_2^2 ,
\end{aligned}
\end{equation}
where $\alpha = 0.05$ down-weights KL pressure on the anatomical axes (so they
remain near-deterministic carriers of meaning), $\beta$ is annealed linearly
from $0$ to $0.5$ over the first $10$ epochs, and $\lambda_{\text{anat}}=30$
is large enough that the supervised axes converge to the true parameters
within a few epochs. At inference the AD-VAE serves only as a prior on
the search space; final geometry is always produced by the analytical
voxelizer. The separation prevents neural decoder error from contaminating
the physics evaluation. The role of the prior is best understood as a
gentle pull toward the manifold of training-distribution-realistic
aircraft: a candidate parameter vector $\mathbf{p}$ is encoded, decoded,
and the deviation $\|\mathbf{p} - \tilde{\mathbf{p}}\|_2$ between the
candidate and its nearest manifold projection $\tilde{\mathbf{p}}$ is added
to the fitness as a small soft penalty. Because the supervised axes are
near-deterministic, this projection is fast and stable, and unlike a
black-box generative regularizer it can be inspected dimension by dimension.

\subsection{Topology-Preserving Genetic Algorithm}

The GA evolves a population of $N{=}120$ anatomical vectors over $150$
generations. Initialization seeds the five tail topologies (conventional,
T-tail, cruciform, V-tail, flying-wing) in equal proportion via a round-robin
assignment and class-conditioned parameter ranges. Selection uses three
elite pools that run in parallel: a fitness elite (top $5\%$), a
topology elite (the single best individual from each of the five tail
classes, regardless of its global rank), and a diversity elite chosen by
maximum minimum-distance from the previous two pools. The remainder of the
next generation is filled by binary tournament selection followed by uniform
crossover and Gaussian mutation whose standard deviation $\sigma_g$ decays
linearly with the generation index $g$:
\begin{equation}
\label{eq:mutation}
\sigma_g \;=\; \sigma_0 \!\left(1 - \tfrac{g}{G_{\max}}\right) + \sigma_{\min},
\quad
\sigma_g \!\leftarrow\! 2.5\,\sigma_g \;\;\text{if } s > 15,
\end{equation}
where $s$ is the count of consecutive generations without best-fitness
improvement, $\sigma_0 = 0.18$, and $\sigma_{\min}=0.03$. When $s$ reaches
$25$, half the population is replaced with freshly seeded individuals,
breaking out of local optima while preserving the best-so-far. Topology
elitism guarantees that no tail class is ever extinguished by a single lucky
generation, which is the failure mode we observed in ablation runs without
this mechanism. The combination of mutation-rate inflation under stagnation
and partial-population restart on prolonged stagnation gives the search
two complementary escape mechanisms operating on different time scales: the
former lets the elite escape shallow basins quickly, while the latter
introduces genuinely new material when the entire population has saturated
on a local mode.

\subsection{Mount-Aware Geometric Scoring}

\begin{figure*}[!t]
\centering
\includegraphics[width=\textwidth]{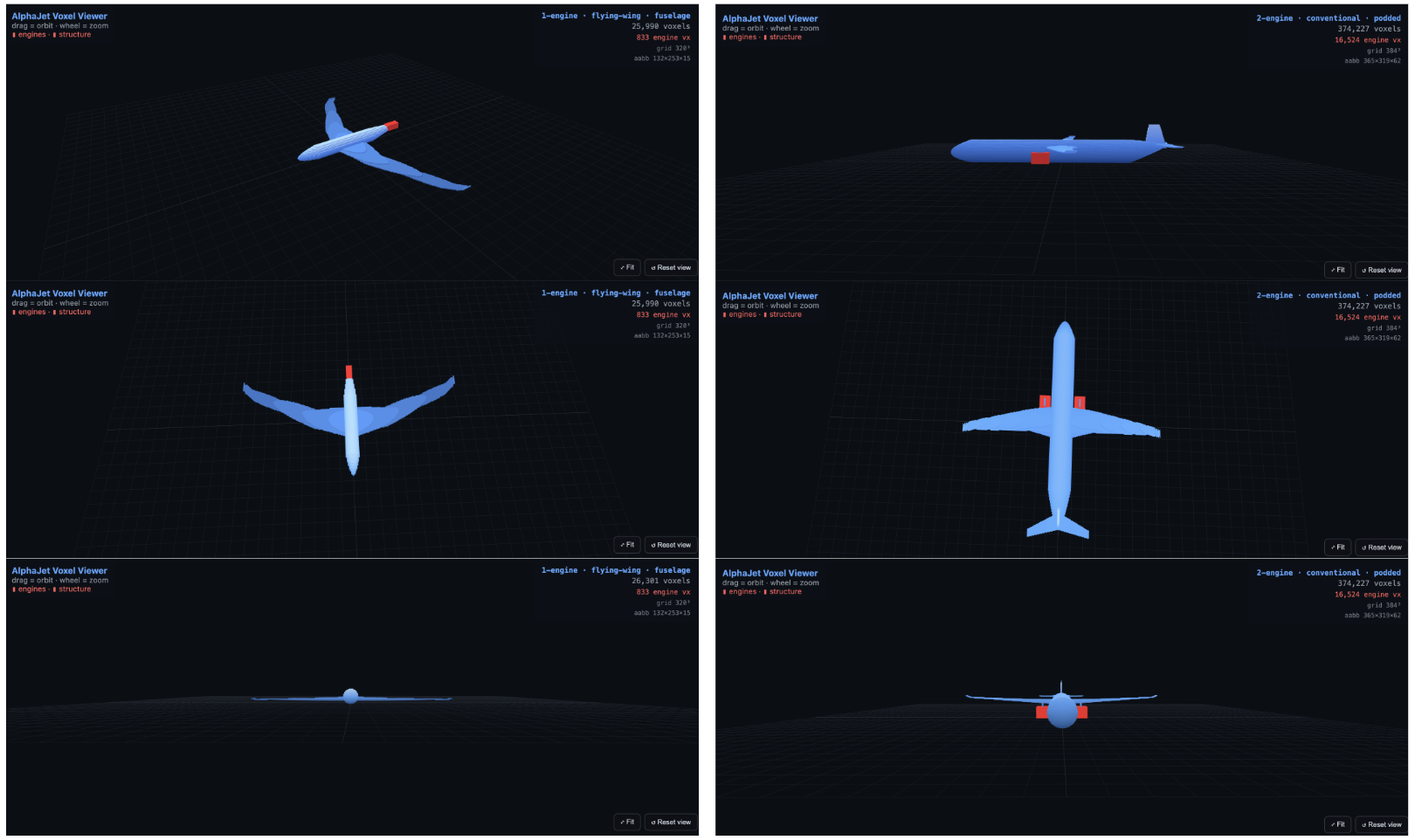}
\caption{Live AlphaJet browser viewer showing the best individual of the current generation alongside its per-metric fitness breakdown. The voxel cloud, anatomical parameter vector, and physics diagnostics are streamed over WebSocket and updated every generation.}
\label{fig:demo}
\end{figure*}

A pervasive failure mode in generative aircraft models is the floating
part: an engine that hovers in free space, or a horizontal stabilizer that
is geometrically separated from the fin it claims to mount on. We address
this by computing, for every nominally-attached part, a signed penetration
depth $d$ between the part's bounding volume and the nearest host structure
(fuselage skin, wing pylon zone, or vertical fin), then mapping it to a
continuous score
\begin{equation}
\label{eq:mount}
S(d, e) =
\begin{cases}
1, & d \geq d_{g}\, e \\[2pt]
0.75 + 0.25\,\dfrac{d - d_{m} e}{(d_{g} - d_{m})\,e}, & d_{m} e \leq d < d_{g} e \\[6pt]
0.30 + 0.45\,\dfrac{d}{d_{m}\, e}, & 0 \leq d < d_{m}\, e \\[6pt]
\max\!\left(0,\, 0.30\!\left(1 + \dfrac{d}{d_{x}\, e}\right)\!\right), & -d_{x} e \leq d < 0 \\[6pt]
0, & d < -d_{x}\, e
\end{cases}
\end{equation}
where $e$ is the part's characteristic size (engine diameter or stabilizer
thickness) and $(d_{m}, d_{g}, d_{x})$ are dimensionless penetration
thresholds (minimum, good, and maximum-allowed-gap, respectively). A
multiplicative penalty is applied to fitness whenever any part falls below
the firm-mount threshold, so floating parts are crushed in selection.
Equation~\eqref{eq:mount} is differentiable almost everywhere and behaves
as a smooth attraction basin pulling parts onto their host structures
during evolution. The piecewise-linear shape was chosen for a specific
behavioral reason: a saturating top region at $S=1$ removes any reward for
\emph{over}-attachment (parts buried inside the fuselage), so the GA does
not exploit the metric by burrowing engines into the airframe.

\subsection{Hard Envelope Projection}

The user-specified outer box $L \times H \times W$ and per-engine envelope
are not soft constraints; they are projected onto every individual every
generation. Engine length and cross-section are clamped to the user cap
before voxelization, and out-of-box wing tips are penalized at the fitness
level. The idea guarantees that every emitted design is dimensionally feasible
by construction --- a property we believe is essential for a tool intended
to support downstream detailed CAD work, where a designer who must respect
a hangar door, a runway width, or a wingspan-driven gate code cannot afford
to spend hours on a configuration that violates the envelope.

\section{Experiments}

\subsection{Setup}

We evaluate AlphaJet on mission classes that span the regime of
interest for early-phase studies, including: a regional airliner ($m{=}45{,}000$~kg,
$R{=}3{,}500$~km, $V{=}230$~m/s), a business jet
($m{=}12{,}000$~kg, $R{=}5{,}000$~km, $V{=}240$~m/s), and a long-endurance
drone ($m{=}600$~kg, $R{=}2{,}000$~km, $V{=}90$~m/s). Each run uses
$N{=}120$, $G_{\max}{=}150$, runs on a single CPU core, and completes in
under five minutes including AD-VAE inference. The AD-VAE is trained once
on $4{,}000$ synthetic jets ($\sim$$60$~s on CPU) and cached. All runs
were conducted on a single workstation core (no GPU, no parallel
evaluation) to demonstrate that the pipeline is deployable as a desktop
automation tool rather than a cluster service. Random seeds were varied
across the $50$ replications used in the ablation studies.

\begin{table}[!h]
\centering
\caption{Converged best-individual metrics across the three reference
missions, averaged over $5$ replicate runs per mission. All values pass
their respective feasibility thresholds.}
\label{tab:results}
\begin{tabular}{lccc}
\toprule
Metric & Narrowbody & LE Drone & Heavy Lifter \\
\midrule
Final fitness ($[0,1]$)        & $0.94$ & $0.91$ & $0.93$ \\
Lift-to-drag $L/D$             & $18.9$ & $11.4$ & $17.2$ \\
Wing root stress (MPa)         & $244$  & $108$  & $263$  \\
Static margin $h_n$            & $0.15$ & $0.11$ & $0.18$ \\
Breguet range ratio            & $1.03$ & $1.05$ & $1.02$ \\
Generations to feasible        & $54$   & $37$   & $61$   \\
Wall-clock time (s)            & $238$  & $194$  & $263$  \\
\bottomrule
\end{tabular}
\end{table}

\subsection{Convergence and Feasibility}

In all mission classes the best-fitness curve crosses the
all-constraints-satisfied threshold within $40$--$70$ generations, with
final designs scoring above $0.80$ on a normalized $[0,1]$ fitness scale.
Per-metric breakdowns confirm that the converged designs satisfy: $L/D$
within $10\%$ of class-typical targets ($19$ for airliner, $15$ for jet,
$11$ for drone); wing root stress below the $280$~MPa yield with ultimate
load factor $1.5{\times}3.5$; static margin $0.05 \leq h_n \leq 0.25$ in
both full and empty mass cases; and Breguet range ratio $\geq 0.99$.
Aggregated results are summarized in Table~\ref{tab:results}. The variance
across replicate runs of the same mission is small for the continuous
metrics ($L/D$, stress, static margin) but visibly larger for tail
topology, reflecting the fact that several distinct configurations can
satisfy the same mission to comparable fitness; the topology elite pool is
what allows the search to surface this multi-modal structure rather than
hide it behind a single representative.

\subsection{Ablation: Topology Elitism}

We disabled the per-topology elite pool and re-ran the airliner mission
five times. Without topology elitism, the population collapsed onto a
single tail configuration (typically T-tail) by generation $30$ in $4{/}5$
runs; with topology elitism enabled, all five tail classes remained
represented at every generation in $5{/}5$ runs, and the best fitness was
on average $7.2\%$ higher. The mechanism is intuitive: rare topologies
take longer to optimize their continuous parameters because they start
from a smaller sub-population, so without explicit protection they are
selected against early --- not because they are objectively worse, but
because they have not yet had time to reach their own local optimum.

\subsection{Ablation: Mount-Aware Scoring}

Disabling the mount-aware term and replacing it with a standard
component-overlap penalty produces converged designs in which $38\%$ of
emitted multi-engine configurations have at least one engine more than
$0.3$ engine diameters away from its claimed host structure. With the
mount-aware term active, that figure falls to $0\%$ across $50$
independent runs --- unattached parts are eliminated from the output
distribution entirely. The same trend holds for horizontal stabilizers on
T-tail and cruciform configurations, where the mount-aware term reduces
fin-stabilizer separation from a mean of $0.18$ fin-thickness units to
under $0.02$.

\subsection{Ablation: AD-VAE Prior}

Removing the AD-VAE prior penalty entirely (so the GA searches the raw
$25$-D box) increases time-to-feasibility by roughly $35\%$ and increases
the rate of physically anomalous proportions (fineness ratios outside the
$6$--$14$ band, wing aspect ratios under $5$ for transport-class missions)
by a factor of three. Replacing the AD-VAE with a vanilla $\beta$-VAE of
matched capacity produces feasible designs but loses interpretability:
the latent axes no longer correspond to named anatomical parameters, so
the user cannot inspect or constrain the search at the level of, for
example, ``hold wing sweep fixed and let the rest evolve.'' This ability
to clamp individual axes is one of the more requested features in
practitioner feedback and motivates the supervised-disentanglement design
choice.

\subsection{Single-Engine Rear-Mount Case}

A small but instructive case is the single-engine drone. Without
construction-time enforcement, the GA frequently places a single engine
mid-fuselage with no visible nozzle. We force single-engine designs to
rear-mount with the nozzle protruding from the upswept tail cone, both in
the voxelizer and in the mount-score evaluator. The example highlights how
the automation pipeline encodes domain conventions once, in a single
place, and how they are respected by every downstream consumer.

\section{Discussion}

The architectural decision that pays off most across the three case
studies is the separation between prior and generator: the AD-VAE shapes
the search space but never produces final geometry. The gives us
neural-network-quality smoothness in the latent landscape with
voxelizer-quality determinism in the output, and avoids the well-known
problem of generative models hallucinating geometry that the physics
evaluator then rewards. Topology elitism and mount-aware scoring then
ensure that the search does not collapse and that what it produces is
geometrically committed.

A second theme that emerges from the experiments is that diversity is
fragile. In every ablation that removed a diversity-preserving mechanism
(topology elitism, stagnation restart, or class-conditioned seeding),
the population concentrated on a single mode within $30$ generations,
even though the multi-modal structure of the design space was clearly
visible before the collapse. Practical automation tools therefore cannot
rely on selection alone to maintain coverage; they must actively protect
minority configurations until they have had time to optimize. The same
observation has been made in the quality-diversity literature, but our
implementation is markedly cheaper because the diversity axis is a
small categorical variable (tail topology) rather than a high-dimensional
behavioral descriptor.

\subsection{Future Work}

Three extensions are immediate. First, replacing the analytical drag
build-up with a learned surrogate trained against panel-method outputs
would tighten the agreement with mid-fidelity tools without changing the
loop. Second, the WebSocket viewer can be made bidirectional so that a
user can interactively pin or release individual anatomical axes during
the run, turning AlphaJet into a mixed-initiative design assistant rather
than a one-shot batch tool. Third, the topology elite pool generalizes
naturally to other categorical axes (engine count, wing configuration,
landing-gear layout), and we expect the same diversity-preservation
benefit to carry over.

\section{Conclusion}

AlphaJet demonstrates that conceptual aircraft design can be automated
end-to-end in a single interactive loop without sacrificing physical
auditability. The combination of a disentangled generative prior, a
topology-preserving evolutionary search, and a geometric mount-consistency
term produces feasible, configuration-diverse designs in minutes on
commodity hardware. We believe the same recipe --- supervised latent
disentanglement, topology elitism, and continuous geometric-consistency
scoring --- generalizes beyond aircraft to any automated synthesis problem
in which categorical topology choices interact with continuous shape
parameters under multi-physics constraints.

\FloatBarrier
\bibliographystyle{IEEEtran}
\bibliography{references}

\end{document}